# Sample Complexity of Multi-task Reinforcement Learning

**Emma Brunskill**
Computer Science Department
Carnegie Mellon University
Pittsburgh, PA 15213

**Lihong Li**
Microsoft Research
One Microsoft Way
Redmond, WA 98052

## Abstract

Transferring knowledge across a sequence of reinforcement-learning tasks is challenging, and has a number of important applications. Though there is encouraging empirical evidence that transfer can improve performance in subsequent reinforcement-learning tasks, there has been very little theoretical analysis. In this paper, we introduce a new multi-task algorithm for a sequence of reinforcement-learning tasks when each task is sampled independently from (an unknown) distribution over a finite set of Markov decision processes whose parameters are initially unknown. For this setting, we prove under certain assumptions that the per-task sample complexity of exploration is reduced significantly due to transfer compared to standard single-task algorithms. Our multi-task algorithm also has the desired characteristic that it is guaranteed not to exhibit negative transfer: in the worst case its per-task sample complexity is comparable to the corresponding single-task algorithm.

## 1 INTRODUCTION

A dream of artificial intelligence is to have lifelong learning agents that learn from prior experience to improve their performance on future tasks. Our interest in the present paper is in how to transfer knowledge and improve performance across a sequence of reinforcement-learning [Sutton and Barto, 1998] problems, where each task itself involves sequential decision making under uncertainty in an unknown environment. We assume that each task is drawn from a finite set of Markov decision processes with identical state and action spaces, but different reward and/or transition model parameters; however, the MDP parameters are initially unknown and the MDP identity of each new task is also unknown. This model is sufficiently rich to capture important applications like tutoring systems that teach a series of students whose initially unknown learning dynamics can be captured by a small set of types (such as honors, standard and remedial), marketing systems that may characterize a customer into a finite set of types and use that to adaptively provide targeted advertising over time, and medical decision support systems that seek to provide good care to patients suffering from the same condition for whom the best treatment strategy may be characterized by a discrete hidden latent variable that captures the patient's physiology.

Although there is encouraging empirical evidence that transferring information across tasks can improve reinforcement learning performance (see Taylor and Stone [2009] for a recent survey), there has been almost no theoretical work to justify or quantify the benefits. This is highlighted as one of the key limitations of the existing research by Taylor and Stone [2009], and there have been only a few papers since then that provide any theoretical analysis [Lazaric and Restelli, 2011, Mann and Choe, 2012]. In particular, we are aware of no work that seeks to formally analyze how transferred knowledge can accelerate reinforcement learning in a *multi-task* settings.

In contrast, there has been a substantial amount of interest over the last decade on Probably Approximately Correct (PAC) reinforcement learning in the single-task setting (e.g. [Kearns and Singh, 2002, Brafman and Tennenholtz, 2002]). This line of work formally quantifies the worst-case learning speed of a reinforcement-learning algorithm, defined as the number of steps in which the agent may fail to follow an $\epsilon$-optimal policy.

In this paper, we introduce a new algorithm for multi-task reinforcement learning, and prove under certain assumptions that the per-task sample complexity is significantly reduced due to transfer compared to the single-task sample complexity. Furthermore, unlike most prior multi-task or transfer reinforcement learning algorithms, our proposed algorithm is guaranteed to avoid negative transfer: the decrease in performance that can arise when misleading information is transferred from a source to target task.

## 2 PRELIMINARIES

This paper focuses on discrete-time, finite Markov decision processes (MDPs) defined as a tuple $\langle S, A, P, R, \gamma \rangle$, where $S$ is the finite state space, $A$ the finite action space, $P$ is the transition probability function, $R$ the reward function, and $\gamma \in (0, 1)$ the discount factor. The reward function is bounded and without loss of generality takes values in $[0, 1]$. For convenience, we also use $S$ and $A$ to denote the cardinality of the state and action spaces, respectively.

A deterministic policy $\pi : S \to A$ defines what action to take in a given state. Its value function, $V^\pi(s)$, is defined as the expected total discounted reward received by executing $\pi$ starting from state $s \in S$. Similarly, the state–action value function, $Q^\pi(s, a)$, is defined as the expected total discounted reward received by taking action $a$ in state $s$ and following $\pi$ thereafter. It is known [Puterman, 1994] that there exist optimal value functions satisfying: $V^* = \max_\pi V^\pi$ and $Q^* = \max_\pi Q^\pi$; furthermore, the greedy policy with respective to $Q^*$ is optimal.

Typically, a reinforcement-learning (RL) [Sutton and Barto, 1998] agent does not know the transition probability and reward functions, and aims to optimize its policy via interaction with the MDP. The main objective of RL is to approximate an optimal policy with as few interactions as possible. One formal framework for analyzing the speed of learning in RL, which we adopt here, is the *sample complexity of exploration* [Kakade, 2003], or *sample complexity* for short. Fix parameters $\epsilon > 0$ and $\delta > 0$. An RL algorithm **A** can be viewed as a nonstationary policy whose value functions can be defined similarly to stationary policies $\pi$ above. At any timestep $h$, we compare the policy value to the optimal policy value in the current state $s_h$. If $V^*(s_h) - V^{\mathbf{A}_h} > \epsilon$, then **A** is not near-optimal in timestep $h$, and a mistake happens. If, with probability at least $1 - \delta$, the total number of mistakes made by an algorithm is at most $\zeta(\epsilon, \delta)$, then $\zeta$ is called the sample complexity of exploration. RL algorithms with a polynomial sample complexity is called PAC-MDP. Further details and related works are found in the survey of Strehl et al. [2009].

In this paper, we consider multi-task RL across a series of $T$ reinforcement-learning tasks, each run for $H$ steps. We assume each task is sampled from a set $\mathcal{M}$ of $C$ MDPs, which share the same state and action spaces, and discount factor, but have different reward and/or transition dynamics. Finally, we denote by $V_{\max}$ an upper bound of the value function. Note that $V_{\max} \leq 1/(1 - \gamma)$, but can be much smaller than this upper bound in many problems.

## 3 PAC-MDP MULTI-TASK RL

We are interested in exploring whether it is possible to reduce sample complexity when the agent faces a sequences of tasks drawn i.i.d. from a distribution, and if so, how this

---

**Algorithm 1** Multi-task RL Algorithm
0: **Input:** $T_1, \bar{C}$.
1: **for** $t = 1, 2, \ldots, T_1$ **do**
2:     Receives an unknown MDP $M_t \in \mathcal{M}$
3:     Run $E^3$ in $M_t$ for $H$ steps to get counts $o(s, a, s', t)$
4: **end for**
5: Combine counts into $\hat{C} \leq \bar{C}$ groups where $\bar{o}(s, a, s', c)$ is the counts for the $c$-th group.
6: **for** $t = T_1 + 1, \ldots, T$ **do**
7:     Receive unknown $M_t \in \mathcal{M}$.
8:     Run Finite-Model-RL on $M_t$
9:     **if** MDP group of task $M_t$ is identified **then**
10:        Incorporate state–action visitation counts from $M_t$ to the group.
11:    **end if**
12: **end for**

---

benefit can be achieved by an algorithm.

Prior work suggests that higher performance is achievable when there is some known structure about the RL MDP parameters. In particular, past research has shown that when a task is drawn from a known distribution over a known finite set MDPs, the problem can be cast as a partially observable MDP planning problem, and solved to yield the Bayes optimal solution if the set cardinality is small [Poupart et al., 2006, Brunskill, 2012]. Although the past work did not examine the sample complexity of this setting, it does suggest the possibility of significant improvements when this structure can be leveraged.

Encouraged by this work, we introduce a two-phase multi-task RL algorithm. Since at the beginning the agent does not know the model parameters, it does single-task learning, and uses the observed transitions and rewards to estimate the parameters of the set of underlying MDPs at the end of phase one. In the second phase, the agent uses these learned models to "accelerate" learning in each new task. We will shortly provide details about both phases of this algorithm, whose performance is formally analyzed in the next section. Before doing so, we also note that our multi-task RL algorithm is designed to minimize or eliminate the potential of negative transfer: tasks where the algorithm performs much worse than a single-task RL algorithm.

Compared to Bayesian approaches, our algorithm development is motivated and guided by sample-complexity analysis. In addition to the guard against negative transfer, our approach is robust, as guaranteed by the theory; this benefit can be shown empirically even in a toy example.

In the following discussion, for clarity we first present a slightly simplified version of our approach (Algorithm 1), before discussing a few additional details in Section 3.3 that involve subtle technicalities required in the analysis.

**Algorithm 2** Finite-Model-RL

0: **Input:** $S, A, \bar{o}, \hat{C}, m, \xi, \epsilon$.
1: Initialize the version space: $\mathcal{C} \leftarrow \{1, \ldots, \hat{C}\}$.
2: $\forall 1 \leq i < j \leq \hat{C}: c_{ij} \leftarrow 0, \Delta_{ij} \leftarrow 0$,
3: $\forall s, a : o(s, a) \leftarrow 0$ (Initialize counts in current task)
4: KNOWN ← Check-Known($S, A, \epsilon, \bar{o}, \mathcal{C}, o, m$)
5: Use $E^3$ algorithm to compute an explore-or-exploit policy and the corresponding value function.
6: Initialize start state $s$
7: **for** $h = 1, \ldots, H$ **do**
8:    Take action $a$, receive reward $r$, and transition to the next state $s'$
9:    **for all** $c \in \mathcal{C}$ **do**
10:      Predict the model dynamics by empirical means: $\hat{\theta}_c \leftarrow \langle \hat{p}(s_1|s,a,c), \ldots, \hat{p}(s_{|S|}|s,a,c), \hat{r}_c \rangle$
11:      Compute the $\ell_2$-confidence interval of $\hat{\theta}_c$ (by, say, Lemma 5); denote the confidence interval by $\delta\theta_c$.
12:    **end for**
13:    Encode the transition $(s, a, r, s')$ by a vector (where **I** is the indicator function)
$z \leftarrow \langle \mathbf{I}(1 = s'), \mathbf{I}(2 = s'), \ldots \mathbf{I}(|S| = s'), r \rangle$
14:    **for all** $i, j \in \mathcal{C}$ such that $i < j$ and $\left\|\hat{\theta}_i - \hat{\theta}_j\right\| \geq 8 \max_{c \in \mathcal{C}} \delta\theta_c$ **do**
15:      $c_{ij} \leftarrow c_{ij} + \frac{1}{4}\left\|\hat{\theta}_i - \hat{\theta}_j\right\|^2$
16:      $\Delta_{ij} \leftarrow \Delta_{ij} + \left\|\hat{\theta}_i - z\right\|^2 - \left\|\hat{\theta}_j - z\right\|^2$
17:      **if** $c_{ij} \geq \xi$ **then**
18:        $\mathcal{C} \leftarrow \mathcal{C} \setminus \{c\}$ where $c = i$ if $\Delta_{ij} > 0$ and $c = j$ otherwise.
19:      **end if**
20:    **end for**
21:    KNOWN ← Check-Known($S, A, \epsilon, \bar{o}, \mathcal{C}, o, m$)
22:    If KNOWN has changed, use $E^3$ to re-compute the policy.
23: **end for**

**Algorithm 3** Check-Known

0: **Input:** $S, A, \epsilon, \bar{o}, \mathcal{C}, o, m$.
1: **for** $(s, a) \in S \times A$ **do**
2:    **if** All MDPs in $\mathcal{C}$ have $\epsilon$-close (in $\ell_2$-norm) estimates of the transition and reward functions at $(s, a)$ **then**
3:      KNOWN$(s, a) \leftarrow$ true
4:    **else if** $o(s, a) \geq m$ **then**
5:      KNOWN$(s, a) \leftarrow$ true
6:    **else**
7:      KNOWN$(s, a) \leftarrow$ false
8:    **end if**
9: **end for**

### 3.1 PHASE ONE

In the first phase, $T_1$ tasks are drawn i.i.d. from the underlying unknown distribution. On each task, the agent follows the single-task algorithm $E^3$ [Kearns and Singh, 2002]. All observed transitions and rewards are stored for each task.

At the end of the first phase, this data is clustered to identify a set of at most $\bar{C}$ MDPs. To do this, the transition and reward parameters are estimated by the empirical means for each task, and tasks whose parameters differ by no more than a fixed threshold are clustered together. After this clustering completes, all the observed transitions and rewards for all tasks in the same cluster are *merged* to yield a single set of data. Our analysis below shows that, under certain assumptions, tasks corresponding to the same MDP will be grouped correctly.

### 3.2 PHASE TWO

At the start of phase two, the agent now has access to a set of (at most) $\bar{C}$ MDPs which approximate the true set of MDP models from which each new task is sampled. The key insight is that the agent can use these candidate models to identify the model of the current task, and then act according to the policy of identified model, and that this process of model identification is generally faster than standard exploration needed in single-task learning.

To accomplish this, we introduce a new single-task RL algorithm, Finite-Model-RL (Algorithm 2), that draws upon but extends the noisy-union algorithm of Li et al. [2011]. One critical distinction is that our approach can be used to compare models that themselves do not have perfect estimates of their own parameters, and do so in way that allows us to eliminate models that are sufficiently unlikely to have generated the observed data.

Like many single-task PAC-MDP RL algorithms, Finite-Model-RL partitions all state–action pairs into known and unknown, where a known state–action pair is one for which we have an $\epsilon$-accurate estimate of its parameters. Following $E^3$, Finite-Model-RL maintains two MDP models for the present task. In the *exploration MDP*, the algorithm assigns unity rewards to unknown states and zero rewards to others. This MDP will be useful for computing a policy to explore unknown states. The other MDP, called the *exploitation MDP*, is identical to the underlying MDP except for unknown states, where rewards are all zero and state transitions are self-loops. This MDP is used to exploit existing knowledge about the current MDP in order to follow a reward-maximization policy.

Similar to $E^3$, our algorithm prioritizes exploration over exploitation: if the optimal value function of the current state in the exploration MDP is above a threshold, the estimated exploration MDP's optimal policy is followed; otherwise, the estimated exploitation MDP's optimal policy is used. In addition, Finite-Model-RL tracks which of the possible set of $\hat{C}$ MDP models could be the underlying MDP of the current task. It eliminates a model when there is sufficient evidence in the observed transitions that it is not the true model (see lines 14–21 in Algorithm 2). We do this by tracking the difference in the sum of the $\ell_2$ error between the current task's observed $(s, a, s', r)$ transitions and the transitions predicted given each of the $\hat{C}$ MDP models obtained at the end of phase 1.

A particular state–action pair becomes known when either there are sufficient observations from the current task that the parameters of that state–action pair can be accurately estimated (as in single-task PAC-MDP RL), or if the remaining possible MDP models have $\epsilon$-close estimates of the state–action pair's parameter in question. If the MDP of the current task is identified (only a single model remains possible in the set), then all the observed data counts from that MDP can be merged with the current task observed data counts. This frequently causes all state–action pairs to become immediately known, and then the algorithm switches to exploitation for the remainder of the task. At the end of each task, the underlying MDP will be identified with high probability, and the observed counts from the current task will be added to the counts for that MDP.

Since the base algorithm in Finite-Model-RL is very much like $E^3$, we preserve the standard single-task sample-complexity guarantee. Thus, negative transfer is avoided in any single task in phase 2, [1] compared to single-task $E^3$.

Across tasks, the observations accumulate and the $\hat{C}$ MDP models will eventually have $\epsilon$-accurate estimates of all state–action parameters. Once this occurs, when facing a new task, as soon as the agent identifies the task model out of the $\hat{C}$ candidates, all state–action pairs become known. We will shortly see that the sample complexity for this identification to occur can be much smaller than standard single-task sample complexity bounds.

### 3.3 ADDITIONAL ALGORITHM DETAILS

We now describe a few additional algorithmic details that have been avoided on purpose to make the main ideas clear. In our analysis later, as well as in some practical situations, these technical details are important.

The key additional detail is that the $E^3$ algorithm is run with two different knownness threshold parameters at different stages of the multi-task algorithm: this parameter specifies the accuracy on the parameter estimates required for a state–action pair to be considered known in standard $E^3$. Usually, this parameter is set to $O(V_{\max}^2/(\epsilon^2(1-\gamma)^2))$ so that once a state–action is known, its dynamics can be estimated sufficiently accurately. However, in multi-task RL, we also need to identify task identity in order to facilitate knowledge transfer to benefit future task.

This observation motivates the use of two different values for this parameter. At the beginning of a task, one may want to use a relatively small value just to do a more balanced random walk in the whole state space, with the primary goal to identify the present task by visiting "informative" states. Here, a state is informative if two MDP models have a sufficient disagreement in its reward or transition dynamics; formal details are given in the next section. Only after the identity is known does the algorithm switch to a larger value, on the order of $O(V_{\max}^2/(\epsilon^2(1-\gamma)^2))$, to learn a near-optimal policy. If, on the other hand, the learner chooses the large value as specified in single-task PAC-MDP algorithms, it is possible that the learner does not visit informative states often enough by the end of a task to know its identity, and the samples collected cannot be transferred to benefit solving future tasks.

More precisely, in phase 1, we first execute $E^3$ with knownness threshold $O(\Gamma^{-2})$, where $\Gamma$, to be defined in the next section, measures the model discrepancy between two MDPs in $\mathcal{M}$, and is in general much larger than $\epsilon$. Once $E^3$ has finished its exploration phase (meaning all state–action pairs have $O(\Gamma)$-accurate parameter estimates), we switch to running $E^3$ with the regular threshold of $O(V_{\max}^2/(\epsilon^2(1-\gamma)^2))$. Since $E^3$ performs all exploration before commencing exploitation, and $\epsilon < \Gamma$, the sample complexity of the resulting method stays the same as initially running $E^3$ with an input $\epsilon$ parameter. This ensures that we maintain the single-task sample-complexity guarantees, but also that we gain enough samples of each state–action pair so as to reliably cluster the tasks at the end of phase 1. With the same approach in phase 2, we can ensure that the task will be identified (with high probability).[2]

Finally, we note that information can also be transferred to the current task through tighter optimistic bounds on the value function that shrink as models are eliminated. Briefly, in phase 2, we can compute an upper bound $\bar{Q}_i$ of the state–action values of the $i \in \hat{C}$ MDPs that also accounts for any uncertainty in the model parameters. At each step, the value of each unknown state–action pair $(s, a)$ can then be set to $\max_{i \in \mathcal{C}} \bar{Q}_i(s, a)$. Since this modification does not seem to impact the worst-case sample complexity, for clarity we did not include it in the description of Algorithm 2, although it may lead to practical improvement.

## 4 ANALYSIS

This section provides an analysis of our multi-task RL algorithm. As mentioned in Section 3.3, two values are used to define the knownness threshold in $E^3$. Due to space limitation, some of the proof details are left to a full version.

To simplify exposition, we use $\theta_i$ to denote MDP $i$'s dy-namics including reward and transitions: the model dynamics in state–action $(s, a)$ is denoted as an $(S + 1)$-dimensional vector $\theta_i(\cdot|s, a)$, where the first $S$ components are the transition probabilities to corresponding next states, and the last component the average reward. The model difference between two MDPs, $M_i$ and $M_j$, in state–action $(s, a)$ is defined as $\|\theta_i(\cdot|s, a) - \theta_j(\cdot|s, a)\|$, the $\ell_2$-difference between their transition probabilities and reward in that state–action. Furthermore, we let $N$ be an upper bound on the number of next states in the transition models in all MDPs in $\mathcal{M}$; while $N$ can be as large as $S$, it can often be much smaller in many realistic problems.

---

[1] Up to log factors, as shown in Section 4.

[2] Note that in phase 2, once the MDP identity of the present task is known, the knownness threshold can switch to the larger value, without having to wait until all state–actions visitation counts reach the $O(\Gamma^{-2})$ threshold.

We make the following assumptions in the analysis:

1. Tasks in $\mathcal{M}$ are drawn from an unknown multinomial distribution, and each task has at least $p_{\min} > 0$ task-prior probability;
2. There is a known upper bound $\bar{C}$ on $C = |\mathcal{M}|$, the number of MDPs in our multi-task RL setting;
3. There is a known gap $\Gamma$ of model difference in $\mathcal{M}$; that is, for all $M_i, M_j \in \mathcal{M}$, there exists some $(s, a)$ such that $\|\theta_i(\cdot|s, a) - \theta_j(\cdot|s, a)\| > \Gamma$.
4. There is a known diameter $D$, such that for every MDP in $\mathcal{M}$, any state $s'$ is reachable from any state $s$ in at most $D$ steps *on average*;
5. All tasks are run for $H = \Omega\left(\frac{DSA}{\Gamma^2} \log \frac{T}{\delta}\right)$ steps;

The first assumption essentially ignores extremely rare MDPs. While it is possible to adapt our results to avoid the assumption of $p_{\min}$, we keep it mostly for the sake of simplicity of the exposition. The second assumption says there are not too many different underlying MDPs. In practice, one may choose $C$ to balance flexibility and complexity of multi-task learning. The third assumption says two distinct MDPs in $\mathcal{M}$ must differ by a sufficient amount in their model parameters; otherwise, there would be little need to distinguish them in practice. The fourth assumption about the diameter, introduced by Jaksch et al. [2010], is the major assumption we need in this work. Basically, it ensures that *on average* every state can be reached from other states sufficiently fast. Consequently, it is possible to quickly identify the underlying MDP of a task.

Our main result is the following theorem: the overall sample complexity in solving $T$ tasks is substantially smaller than solving them individually without transfer.

**Theorem 1** *Given any $\epsilon$ and $\delta$, run Algorithm 1 for $T$ tasks, each for $H = \Omega\left(\frac{DSA}{\Gamma^2} \log \frac{T}{\delta}\right)$ steps. Then, the algorithm will follow an $\epsilon$-optimal policy on all but $\tilde{O}\left(\frac{\zeta V_{\max}}{\epsilon(1-\gamma)}\right)$ steps, with probability at least $1 - \delta$, where*

$$\zeta = \tilde{O}\left(T_1 \zeta_s + \bar{C}\zeta_s + (T - T_1)\left(\frac{NV_{\max}^2 \bar{C}}{\epsilon^2(1-\gamma)^2} + \frac{DC^2}{\Gamma^2}\right)\right),$$

*and $\zeta_s = \tilde{O}\left(\frac{NSAV_{\max}^2}{\epsilon^2(1-\gamma)^2}\right)$, with probability at least $1 - \delta$.*

In particular, in phase 2, our Algorithm 1 has a sample complexity that is independent of the size of the state and action spaces, trading this for a dependence on the number of models $\bar{C}$ and diameter $D$. In contrast, applying single-task learning without transfer in $T$ tasks can lead to an overall sample complexity of $\tilde{O}(T\zeta_s) = \tilde{O}(TNSA)$. Since we expect $\bar{C} \ll SA$, this yields a significant improvement over single-task reinforcement learners (as long as $D$ is not too large), whose sample complexity has at least a linear dependence on the size of the state–action space [Strehl et al., 2006b, Szita and Szepesvári, 2010], and some have a polynomial dependence on the size of the state and/or action spaces. We expect this reduction in sample complexity to also lead to improved empirical performance, and verify this in an experiment later.

A few lemmas are needed to prove the main theorem.

**Lemma 1** *If we set $T_1 = p_{\min}^{-1} \ln \bar{C}/\delta$, then with probability $1 - \delta$, all MDPs will be encountered in phase 1.*

Proof. In the $T_1$ samples in phase 1, the probability that every MDP is seen at least once is no smaller than $1 - \bar{C}(1 - p_{\min})^{T_1}$. Setting this lower bound to $1 - \delta$, solving for $T_1$, and using the inequality $\ln(1 - x) < -x$, we get $T_1 = \frac{1}{p_{\min}} \ln \frac{\bar{C}}{\delta}$ is sufficient. □

**Lemma 2** *If all tasks are run for $H = \tilde{O}\left(\frac{DSA}{\Gamma^2}\right)$ steps, then with probability $1 - \delta$, the following hold:*

1. *Every state–action in every task receives at least $\Omega(\Gamma^{-2} \ln T/\delta)$ samples from that task;*
2. *The tasks encountered in phase 1 will be grouped correctly with all other tasks corresponding to the same (hidden) MDP;*
3. *Each task in phase 2 will be identified correctly and its counts added to the correct MDP.*

Proof. Assumption 4 ensures that any state is reachable from any other state within $2D$ steps with probability at least $0.5$, by Markov's inequality. Chernoff's inequality, combined with a union bound over all $T$ tasks and all $SA$ state–action pairs, implies that with probability at least $1 - \delta$, all state–actions can be visited $\Omega(\Gamma^{-2} \ln \frac{TSA}{\delta})$ times as long as sufficiently large $H$.

The second statement is proved by Hoeffding's inequality. After phase 1 each task will have at least $\Omega(\Gamma^{-2} \ln T/\delta)$ samples for each state–action with high probability. In order to accurately merge tasks into groups implicitly associated with the same underlying MDP, we note that by assumption, any two different MDPs must have dynamics that differ by at least $\Gamma$ in at least one state–action. In order to detect such a difference, it is sufficient to estimate the models of each state–action to an $\ell_2$-accuracy of $\Gamma/4$. In this case, the $\ell_2$-difference between any two MDPs must exceed $\Gamma/2$ in at least one state–action pair. A similar analysis of two tasks which come from the same MDP implies that the difference estimated mean rewards can be at most $\Gamma/2$ for all state–actions. This implies that tasks can be clustered into groups corresponding to all tasks from the same MDP by combining tasks whose reward models differ by no more than $\Gamma/2$ across all state–action pairs. This is ensured by Hoeffding's inequality with a union bound, resulting in the sample size of $\Omega\left(\Gamma^{-2}\ln\frac{TSA}{\delta}\right)$.

The third part requires that each tasks's MDP identity can be correctly identified with high probability in phase 2, which can be proved similarly to the second part. □

The next lemma shows, on average, each state transition contains information to distinguish the true MDP model from others. Let $\theta_1$ and $\theta_2$ be two $(S+1)$-dimensional vectors, representing two MDP models for some state–action $(s,a)$. Let $\hat{\theta}_1$ and $\hat{\theta}_2$ be their estimates that have confidence radius $\delta\theta_1$ and $\delta\theta_2$, respectively; that is, $\left\|\theta_1-\hat{\theta}_1\right\|\le\delta\theta_1$ and similar for $\theta_2$. For a transition $(s,a,r,s')$, define the square loss of estimated model $\hat{\theta}_i$ by

$$\ell(\hat{\theta}_i)=\sum_{1\le\tau\le S,\tau\ne s'}\hat{\theta}_i(\tau)^2+(\hat{\theta}_i(s')-1)^2+(\hat{\theta}_i(S+1)-r)^2.$$

**Lemma 3** *If $\theta_1$ is the true model for generating the transition $(s,a,r,s')$, then*

$$\mathbb{E}_{\theta_1}\left[\ell(\hat{\theta}_2)-\ell(\hat{\theta}_1)\right]\ge\left\|\hat{\theta}_1-\hat{\theta}_2\right\|\left(\left\|\hat{\theta}_1-\hat{\theta}_2\right\|-2\left\|\delta\theta_1\right\|\right).$$

Proof. Written out explicitly, the left-hand side becomes

$$\sum_i\Big(\theta_1(i)\left((1-\hat{\theta}_2(i))^2-(1-\hat{\theta}_1(i))^2\right)$$
$$+(1-\theta_1(i))(\hat{\theta}_2(i)^2-\hat{\theta}_1(i)^2)\Big)$$
$$+\mathbb{E}_{r\sim\theta_1}\left[(r-\theta_2(S+1))^2-(r-\theta_1(S+1))^2\right].$$

Some algebra simplifies the above as:

$$\sum_{\tau=1}^{S+1}\left(\hat{\theta}_1(\tau)-\hat{\theta}_2(\tau)\right)\left(2\theta_1(\tau)-\hat{\theta}_1(\tau)-\hat{\theta}_2(\tau)\right)$$
$$=\sum_{\tau=1}^{S+1}\left(\hat{\theta}_1(\tau)-\hat{\theta}_2(\tau)\right)\left(\hat{\theta}_1(\tau)-\hat{\theta}_2(\tau)2\theta_1(\tau)-2\hat{\theta}_1(\tau)\right)$$
$$=\left\|\hat{\theta}_1-\hat{\theta}_2\right\|^2+2\langle\hat{\theta}_1-\hat{\theta}_2,\theta_1-\hat{\theta}_1\rangle$$
$$\ge\left\|\hat{\theta}_1-\hat{\theta}_2\right\|^2-2\left\|\hat{\theta}_1-\hat{\theta}_2\right\|\left\|\theta_1-\hat{\theta}_1\right\|$$
$$\ge\left\|\hat{\theta}_1-\hat{\theta}_2\right\|\left(\left\|\hat{\theta}_1-\hat{\theta}_2\right\|-2\left\|\delta\theta_1\right\|\right),$$

where the first inequality is due to Cauchy inequality, and the second to the condition in the lemma. □

We are going to apply a generic PAC-MDP theorem of Strehl et al. [2006a] to analyze the sample complexity of our algorithm. As usual, define a state–action to be known if its reward estimate is within $\Theta(\epsilon(1-\gamma))$ accuracy, and its next-state transition probability estimate is within $\Theta(\epsilon(1-\gamma)/V_{\max})$ in terms of total variation. The next lemma bounds the number of visits to unknown state–actions in the entire second phase.

**Lemma 4** *The total number of visits to unknown state–actions in the the second phase is*

$$\tilde{O}\left(\frac{(T-T_1)DC^2}{\Gamma^2}+\frac{NV_{\max}^2C(T-T_1)}{\epsilon^2(1-\gamma)^2}\ln\frac{C}{\delta}+C\zeta_s\right).$$

Proof. (sketch) As explained earlier, Algorithm 2 starts with model identification, and then switches to single-task E$^3$. The first term in the bound corresponds to the model identification step. Note that, for a set of $C$ models, there are at most $C$-choose-2, namely $O(C^2)$, many informative states to fully identify a model. Therefore, our algorithm only needs to reach these informative states before figuring out the true model. Similar to the proof of Lemma 2 (part 1), each such state can be visited $\Theta(\Gamma^{-2})$ times in $\tilde{O}(D\Gamma^{-2})$ steps. So, $\tilde{O}(D\Gamma^{-2}C^2)$ steps suffice to visit all these informative states sufficiently often.

The rest of the proof (for the second and third terms) consists of two parts. The first assumes the underlying MDP identity is known at the beginning of each task. In our algorithm, however, the MDP identity is unknown until all but one model is eliminated. Then, the second part shows such a delay of MDP identification is insignificant with respect to the number of visits to unknown state–actions.

We begin with the assumption that the underlying MDP identity is given at the beginning of each task. Although the algorithm knows which MDP it is in in the current task, it still follows the same logic in the pseudocode for model elimination and identification. The only advantage it has is to "boost" its history with samples of previous tasks of the same MDP right after the task begins, rather than at the end of the task. This is like single-task learning by "concatenating" tasks of the same MDP into one big task.

In this scenario, an unknown state–action $(s,a)$ implies at least one of the following must be true. The first is when the number of samples of $(s,a)$ in some model has not exceeded the known threshold $\zeta_s/(SA)$. For this case, since the samples of $(s,a)$ for the same MDP accumulates over tasks, there can be at most $\zeta_s/(SA)$ visits to unknown $(s,a)$ pairs for a single MDP model, and a total of $C\zeta_s/(SA)$ visits to unknown $(s,a)$ across all $C$ models. In the other case, at least two models in $\mathcal{M}$ has a sufficient difference in their estimates of the model parameters for $(s,a)$. Using Lemma 3, one can calculate the expected difference in square loss between the true model and a wrong model. Following similar steps as in [Li et al., 2011],[3] we can see the squared difference on average is at least

---

[3] The noisy union algorithm of [Li et al., 2011] is based on

$\Theta(\epsilon^2(1-\gamma)^2/(V_{\max}^2 N))$, and after $O(\frac{NV_{\max}^2 C}{\epsilon^2(1-\gamma)^2} \ln \frac{CT}{\delta})$ visits to such state–actions, all models but the true one will be eliminated, with probability at least $1 - \frac{\delta}{CT}$. Using a union bound over all tasks in phase 2, we have that, with probability at least $1 - \delta$, the same statement holds for all tasks in phase 2. Details will be given in a full version.

The first part of the proof is now completed, showing that when the task identity is given at the beginning of a task, the total number of visits to unknown state–actions is at most $O\left(\frac{NV_{\max}^2 C(T-T_1)}{\epsilon^2(1-\gamma)^2} \ln \frac{C}{\delta} + C\zeta_s\right)$.

We now handle the need for MDP identification, which prevents samples in the present task to contribute to the corresponding model until the underlying MDP is identified. Consider any state–action pair $(s,a)$, and a fixed task in phase 2. At the beginning of the task, the algorithm has access to $C$ models, the $i$-th of which has accumulated $U_i$ samples for $(s,a)$. After the task, the true MDP (say, model 1, without loss of generality) is identified, whose sample count for $(s,a)$ becomes $U_1' \leftarrow U_1 + U_0$, where $U_0$ is the number of visits to $(s,a)$ in the present task. For other models $i > 1$, $U_i' \leftarrow U_i$.

Consider three situations regarding the sample sizes $U_1$ and $U_1'$. In the first case, $U_1 < U_1' < \zeta_s/(SA)$, so our multi-task RL algorithm behaves identically no matter whether the samples contribute to the true model estimation immediately or at the end of the task, since $(s,a)$ will remain unknown in either situation. In the second case, $\zeta_s/(SA) \leq U_1 < U_1'$, so $(s,a)$ is already know at the beginning of the task, and additional samples for $(s,a)$ does not change the algorithm, or increase the number of visits to unknown state–actions. In the last case, we have $U_1 < \zeta_s/(SA) \leq U_1'$. Recall that our algorithm declares a state–action to be known if it has been visited $\zeta_s/(SA)$ times in a single task, so $U_0 \leq \zeta_s/(SA)$. Hence, $U_1' = U_1 + U_0 < 2\zeta_s/(SA)$. Applying this inequality to all state–actions and all MDPs, we conclude that the number of visits to unknown states is at most $2C\zeta_s$.

Part II above shows the delay in sample accumulation can only cause up to a constant factor increase in the number of visits to unknown state–actions. The lemma follows immediately from the conclusion of part I. □

We are now fully equipped to prove the main result:

Proof. (of Theorem 1) We will use the generic PAC-MDP theorem of Strehl et al. [2006a] by verifying the three needed conditions hold. Although the theorem of [Strehl et al., 2006a] is stated for single-task RL, the proof works without essential changes in multi-task RL.

The first condition holds since the value function is optimistic with high probability, by construction of the known-state MDPs when running E$^3$ in the tasks.

---

scalar predictions and observations, while we are dealing with $(S+1)$-dimensional vectors. The only substantial change to their proof is to replace the application of Hoeffding's inequality with its vector-valued extensions, such as Lemma 5 in the appendix.

| 1 | 2 | 3 | 4 | 5 |
| --- | --- | --- | --- | --- |
| 6 | 7 | 8 | 9 | 10 |
| 11 | 12 | 13 | 14 | 15 |
| 16 | 17 | 18 | 19 | 20 |
| 21 | 22 | 23 | 24 | 25 |

Figure 1: Gridworld domain

The second condition also holds. Whenever a state–action becomes known, its reward estimate is within $\epsilon(1-\gamma)$ accuracy, and the transition estimate is within $\epsilon(1-\gamma)/(V_{\max}\sqrt{N})$ accuracy measured by $\ell_2$ error. Using the inequality $\|v\|_1 \leq \|v\|_2 \sqrt{d}$, where $d$ is the dimension of vector $v$, we know the transition estimate is within $\epsilon(1-\gamma)/V_{\max}$ accuracy measured by total variation. The simulation lemma (see, e.g., [Strehl et al., 2006a]) then implies the accuracy condition holds.

What remains to be shown is that the third condition holds; namely, we will find a bound on the total number of times an unknown state–action pair is visited, across all $T$ tasks. E$^3$ is executed on each task in phase 1. Prior analysis for Rmax and MBIE (see e.g. [Kakade, 2003, Strehl and Littman, 2008]) applies similarly to E$^3$, implying that the number of visits to unknown state–action pairs on a single MDP at most $\left(\frac{SANV_{\max}^2}{\epsilon^2(1-\gamma)^2}\right)$. From Lemma 1, after $T_1$ tasks, all tasks in $\mathcal{C}$ have been encountered with probability at least $1 - \delta$. Therefore, with probability at least $1 - \delta$, the total number of visits to unknown state–action pairs in phase 1 is at most $\zeta_s T_1$.

In Phase 2, Algorithm 1 runs E$^3$ in individual tasks but transfers samples from one task to another of the same underlying MDP. We have shown in Lemma 4 the number of visits to unknown state–actions is at most $O\left(\frac{(T-T_1)DC^2}{\Gamma^2} + \frac{NV_{\max}^2 C(T-T_1)}{\epsilon^2(1-\gamma)^2} \ln \frac{C}{\delta} + C\zeta_s\right)$. Hence, the total number of visits to unknown state–actions during all $T$ tasks is at most $O\left(\zeta_s T_1 + (T-T_1)\left(\frac{DC^2}{\Gamma^2} + \frac{NV_{\max}^2 C}{\epsilon^2(1-\gamma)^2} \ln \frac{C}{\delta}\right) + C\zeta_s\right)$.
The theorem follows immediately by the PAC-MDP theorem of [Strehl et al., 2006a]. □

## 5 EXPERIMENTS

Although the main contribution of our paper is to provide the first theoretical justification for online multi-task Rl, we also provide numerical evidence showing the empirical benefit of our proposed approach over single-task learning as well as a state-of-the-art multi-task algorithm.

There are $C = 3$ possible MDPs, each with the same $5 \times 5$ state space as shown in the gridworld layout of Figure 5. The start state is always the center state ($s_{13}$). There are 4 actions that succeed in generally moving the agent in the intended cardinal direction with probability 0.85, going in the other directions (unless there is a wall) with probability 0.05 each. Three of the corners ($s_5, s_{21}, s_{25}$) exhibit different dynamics: the agent stays in the state with probability 0.95, or otherwise transitions back to the start state. The three MDPs differ only in their reward models. The reward for each state is drawn from a binomial model. Intuitively, in each MDP, one of the corners provides a high reward, two others provide low reward, and there is one additional state whose medium reward can help distinguish the MDPs. More precisely, In MDP 1, $s_{21}$ has a binomial parameter of 0.99, $s_6$ has a parameter of 0.6, $s_5$ and $s_{25}$ have a parameter of 0, and all other states have a parameter of 0.1. In MDP 2, $s_5$ has a binomial parameter of 0.99, $s_2$ has a parameter of 0.6, $s_{21}$ and $s_{25}$ have a parameter of 0, and all other states have a parameter of 0.1. In MDP 3 $s_{25}$ has a binomial parameter of 0.99, $s_1$ has a parameter of 0.6, $s_{21}$ and $s_{25}$ have a parameter of 0, and all other states have a parameter of 0.1. A new task is sampled from one of these three MDPs with equal probability.

We compare our proposed approach to the most closely related approach we are aware of, Wilson et al. [2007]'s algorithm on hierarchical multi-task learning (HMTL). Wilson et al. learn a Bayesian mixture model over a set of MDP classes as the agent acts in a series of MDPs, and use prior to transfer knowledge to a new task sampled from one of those classes. When acting in a new MDP, their approach does not explicitly balance exploration and exploitation; instead, it selects the current maximum a posterior (MAP) estimate of the model parameters, computes a policy for this model, and uses this policy for a fixed number of steps, before re-computing the MAP model. Wilson et al. did not provide formal performance guarantees for their approach, but they did achieve promising results on both a simulated domain and a real-time strategy game with HMTL. When applying their approach to our setting, we limit the hierarchy to one level, ensuring that tasks are directly sampled from a mixture of MDPs. We also provide their algorithm with an upper bound on the number of MDPs, though their algorithm is capable to learning this directly.

In both our algorithm and HMTL there are several parameters to be set. For HMTL we set the interval between recomputing the MAP model parameters at 10 steps: this was chosen after informal experimentation suggested this improved performance compared to longer intervals. In our approach we set the threshold for a parameter to be known at $m = 5$. The number of tasks in phase 1 was set to $\lceil 3 \ln(3/0.05) \rceil = 13$, matching the required length specified by Lemma 1. We ran each task for a horizon of $H = 3000$ steps, and performed multi-task reinforcement

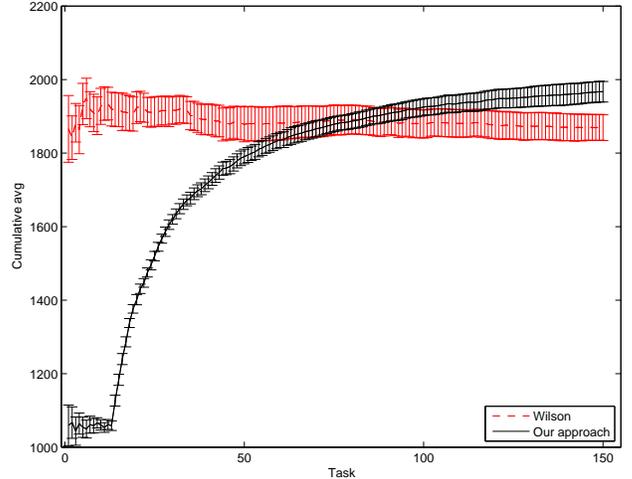

Figure 2: Cumulative average reward per task.

learning across 150 tasks per round. We then repeated this process for 20 rounds.

Figure 5 displays the cumulative per-task reward for each method, averaged across 20 rounds. As expected, our approach performs worse during phase 1, before it has obtained a good estimate of each of the 3 MDPs. In phase 2, our approach performs well, successfully leveraging its knowledge of the models to quickly determine the new task's MDP identity, and then act optimally for that MDP for the remainder of the task. Since phase 1 of our algorithm runs single-task $E^3$, we can see that transferring knowledge enables our approach to perform substantially better than single-task $E^3$ ($p < 10^{-4}$ in a Mann-Whitney U test comparing the first task performance to the last).

HMTL does quite well even at the start, because the algorithm directly exploits the current estimated parameters, and once the agent bumps into a good state, the algorithm can leverage that information for the remainder of the task. However, HMTL does not significantly improve beyond its original performance, and performs similarly across the entire length of a multi-task round. We hypothesize that this is because in each task, this approach is not explicitly performing exploration, and therefore may only get good estimates of the parameters of some of the state–action pairs. This means it can be harder to learn a good estimate of the mixture model over the MDPs. Indeed, when we examine the multi-task posterior learned by HMTL, we find that the resulting MDPs appear to be *mixtures* of the true set of MDPs. We compare the total reward obtained in a single round (of all 150 tasks in both phases) of the two approaches, and our approach achieved significantly higher total reward ($p = 0.03$ in a Mann-Whitney U test).

These results provide empirical evidence that our algorithm both achieves significantly better sample-complexity results than prior single-task algorithms as well as a state-of-the-art multi-task algorithm, and these gains can indeed translate to improved empirical performance.

## 6 RELATED WORK

Our setting most closely matches that of Wilson et al. [2007]; however, they consider a more general two-level hierarchical model where tasks are sampled from a class distribution, and there is a mixture over classes. The authors update a Bayesian prior over the hierarchical model parameters after finishing acting in each task using MCMC, and use and update a local version of this prior during each single-task by sampling an MDP from this prior, following the model's optimal policy for a fixed number of steps, updating the prior, and repeating. Though the authors demonstrate promising empirical learning improvements due to transfer, unlike our work, no formal analysis is provided.

Other work studies the related problem of *transfer RL*. For example, Lazaric and Restelli [2011] provide value-function approximation error bounds of the target task in a *batch* setting, as opposed to our *online* setting where the agent has to balance exploration and exploitation. Their bounds quantify the error potentially introduced by transferring source-task samples to an unrelated target task as well as the reduction in error due to increasing the number of samples from the source. Sorg and Singh [2009] prove bounds on transferring the state–action values from a source MDP to a target MDP, where both MDP models are known and there exists a soft homomorphism between the two state spaces. If the target MDP model is unknown, the authors present a promising heuristic approach without performance guarantees. More recently, Mann and Choe [2012] introduce an algorithm that uses a slight modification of a source task's optimal value as an optimistic initial value for a subset of the target task's state–action pairs, given a mapping between the tasks that ensures the associated value functions have similar values. The authors provide characteristics of this mapping that will improve sample complexity of their algorithm. While interesting, no algorithm was given that could meet these conditions, and no sample-complexity bounds were provided. Perhaps most similar to us is Mehta et al. [2008], who consider sample complexity for transfer learning across semi-MDPs with identical $(S, A)$ and transition models, and a distribution of reward weight vectors. The authors provide a bound on the number of tasks needed until they will be able to immediately identify a close-to-optimal policy for a future task. Compared to our work: (1) the authors only use transferred information to initialize the value function in the new task, (2) their algorithm can produce negative transfer, and more significantly, (3) the authors assume that the model of each new semi-MDP is completely specified.

The model-elimination idea in our Algorithm 2 is related to several previous work on *single-task* RL. The most relevant is probably the (more special) noisy union algorithm of Li et al. [2011] and its application to Met-Rmax [Diuk et al., 2009]. Here, the noisy union algorithm is generalized so that model elimination is possible even before a state–action becomes fully known. Similar ideas are also found in the Parameter Elimination algorithm [Dyagilev et al., 2008], which uses Wald's Sequential Probability Ratio Test (SPRT) to eliminate models, as opposed to the simpler square loss metric we use here. Finally, Lattimore et al. [2013] employ model elimination in their Maximum Exploration algorithm that works in general reinforcement-learning problems beyond MDPs.

## 7 CONCLUSIONS

In this paper, we analyze the sample complexity of exploration for a multi-task reinforcement-learning algorithm, and show substantial advantage compared to single-task learning. In contrast to the majority of the literature, this work is theoretically grounded, using tools in the PAC-MDP framework. Furthermore, we also show the possibility of avoiding negative transfer in multi-task RL.

These promising results suggest several interesting directions for future research. One of them is to relax some of the assumptions and to develop more broadly applicable algorithms. Second, we intend to test the proposed algorithm in benchmark problems and investigate its empirical advantage compared to single-task RL, as well as its robustness with respect to parameters like $\bar{C}$. Third, it is interesting to extend the current results beyond finite MDPs, possibly relying on function approximation or compact model representations like dynamic Bayes networks [Dean and Kanazawa, 1989]. Finally, our algorithm makes use only of the learned MDP parameters, not of the task distribution over $\mathcal{M}$. Although our own attempt has not yet identified theoretical benefits from such information, we suspect at the least that it will be empirically beneficial.

## A A CONCENTRATION INEQUALITY

The following result extends Hoeffding's inequality from real-valued random variables to vector-valued random variables. The tail probability upper bound here is only a constant factor worse than that of Hoeffding's.

**Lemma 5 (Hayes [2005])** *For vector-valued martingale,* $\Pr\left(\|X_n\| \geq a\right) \leq 2\exp\left(2 - \frac{a^2}{2n}\right);$ *or equivalently,* $\Pr\left(\left\|\frac{X_n}{n}\right\| \geq \epsilon\right) \leq 2\exp\left(2 - \frac{n\epsilon^2}{2}\right).$

**Acknowledgements**

We thank Chris Meek for a helpful discussion of the general differences between transfer learning and multi-task learning. E.B. also thanks Google for partial funding support.

# References

R. I. Brafman and M. Tennenholtz. R-max—a general polynomial time algorithm for near-optimal reinforcement learning. *Journal of Machine Learning Research*, 3:213–231, October 2002.

E. Brunskill. Bayes-optimal reinforcement learning for discrete uncertainty domains. In *Proceedings of the Eleventh International Conference on Autonomous Agents and Multiagent Systems (AAMAS)*, pages 1385–1386, 2012.

T. Dean and K. Kanazawa. A model for reasoning about persistence and causation. *Computational Intelligence*, 5(3):142–150, 1989.

C. Diuk, L. Li, and B. R. Leffler. The adaptive $k$-meteorologists problem and its application to structure discovery and feature selection in reinforcement learning. In *Proceedings of the Twenty-Sixth International Conference on Machine Learning (ICML)*, pages 249–256, 2009.

K. Dyagilev, S. Mannor, and N. Shimkin. Efficient reinforcement learning in parameterized models: Discrete parameter case. In *Recent Advances in Reinforcement Learning*, volume 5323 of *Lecture Notes in Computer Science*, pages 41–54, 2008.

T. P. Hayes. A large-deviation inequality for vector-valued martingales, 2005. Unpublished manuscript.

T. Jaksch, R. Ortner, and P. Auer. Near-optimal regret bounds for reinforcement learning. *Journal of Machine Learning Research*, 11:1563–1600, 2010.

S. M. Kakade. *On the Sample Complexity of Reinforcement Learning.* PhD thesis, University College London, 2003.

M. J. Kearns and S. P. Singh. Near-optimal reinforcement learning in polynomial time. *Machine Learning*, 49(2–3):209–232, 2002.

T. Lattimore, M. Hutter, and P. Sunehag. The sample-complexity of general reinforcement learning. In *Proceedings of Thirtieth International Conference on Machine Learning (ICML)*, 2013. To appear.

A. Lazaric and M. Restelli. Transfer from multiple MDPs. In *Proceedings of the Neural Information Processing Systems (NIPS)*, pages 1746–1754, 2011.

L. Li, M. L. Littman, T. J. Walsh, and A. L. Strehl. Knows what it knows: A framework for self-aware learning. *Machine Learning*, 82(3):399–443, 2011.

T. A. Mann and Y. Choe. Directed exploration in reinforcement learning with transferred knowledge. In *European Workshop on Reinforcement Learning*, 2012.

N. Mehta, S. Natarajan, P. Tadepalli, and A. Fern. Transfer in variable-reward hierarchical reinforcement learning. *Machine Learning*, 73(3):289–312, 2008.

P. Poupart, N. Vlassis, J. Hoey, and K. Regan. An analytic solution to discrete Bayesian reinforcement learning. In *Proceedings of the International Conference on Machine Learning (ICML)*, pages 697–704, 2006.

M. L. Puterman. *Markov Decision Processes: Discrete Stochastic Dynamic Programming*. Wiley-Interscience, New York, 1994. ISBN 0-471-61977-9.

J. Sorg and S. P. Singh. Transfer via soft homomorphisms. In *Proceedings of the 8th International Conference on Autonomous Agents and Multiagent System (AAMAS)*, pages 741–748, 2009.

A. L. Strehl and M. L. Littman. An analysis of model-based interval estimation for Markov decision processes. *Journal of Computer and System Sciences*, 74(8):1309–1331, 2008.

A. L. Strehl, L. Li, and M. L. Littman. Incremental model-based learners with formal learning-time guarantees. In *Proceedings of the Twenty-Second Conference on Uncertainty in Artificial Intelligence (UAI)*, pages 485–493, 2006a.

A. L. Strehl, L. Li, E. Wiewiora, J. Langford, and M. L. Littman. PAC model-free reinforcement learning. In *Proceedings of the Twenty-Third International Conference on Machine Learning (ICML)*, pages 881–888, 2006b.

A. L. Strehl, L. Li, and M. L. Littman. Reinforcement learning in finite MDPs: PAC analysis. *Journal of Machine Learning Research*, 10:2413–2444, 2009.

R. S. Sutton and A. G. Barto. *Reinforcement Learning: An Introduction*. MIT Press, Cambridge, MA, March 1998. ISBN 0-262-19398-1.

I. Szita and C. Szepesvári. Model-based reinforcement learning with nearly tight exploration complexity bounds. In *Proceedings of the Twenty-Seventh International Conference on Machine Learning (ICML)*, pages 1031–1038, 2010.

M. E. Taylor and P. Stone. Transfer learning for reinforcement learning domains: A survey. *Journal of Machine Learning Research*, 10(1):1633–1685, 2009.

A. Wilson, A. Fern, S. Ray, and P. Tadepalli. Multi-task reinforcement learning: a hierarchical Bayesian approach. In *Proceedings of the International Conference on Machine Learning (ICML)*, pages 1015–1022, 2007.